\documentclass[10pt,twocolumn,letterpaper]{article}

\usepackage{iccv}
\usepackage{times}
\usepackage{epsfig}
\usepackage{graphicx}
\usepackage{amsmath}
\usepackage{amssymb}

\usepackage{multirow}
\usepackage{booktabs}
\usepackage{pifont}

\usepackage[abs]{overpic}
\usepackage{pict2e}

\usepackage[table, dvipsnames]{xcolor}

\usepackage{csquotes}

\newcommand{\R}{\mathbb{R}}
\newcommand{\E}{\mathbb{E}}
\newcommand{\p}{\text{p}}

\newcommand{\SOT}{\text{SO(3)}}
\newcommand{\SET}{\text{SE(3)}}

\newcommand{\Ours}{Ours}
\newcommand{\OursNoKP}{\Ours\ w/o KP}
\newcommand{\OursNoIS}{\Ours\ w/o IS}
\newcommand{\OursNoKPIS}{\Ours\ w/o KP \& IS}
\newcommand{\figspacing}{\vspace{0em}}

\usepackage[breaklinks=true,bookmarks=false,hidelinks=false]{hyperref}

\iccvfinalcopy % *** Uncomment this line for the final submission

\def\iccvPaperID{1} % *** Enter the ICCV Paper ID here

\begin{document}

%%%%%%%%% TITLE
\title{
SpyroPose: SE(3) Pyramids for Object Pose Distribution Estimation
%Importance Sampling Pyramids \\for Object Pose Distribution Estimation in SE(3)
}
%evt. Importance Sampling Pyramids for Object Pose Distribution estimation in SE(3)
%\\with Contrastive Learning and Importance Sampling}

\author{Rasmus Laurvig Haugaard\\
\and
Frederik Hagelskjær\\
\and 
Thorbjørn Mosekjær Iversen\\
\and 
\vspace{-2em}\\
SDU Robotics, University of Southern Denmark\\
{\tt\small \{rlha,frhag,thmi\}@mmmi.sdu.dk}
}
\maketitle
% \ificcvfinal\pagestyle{empty}\thispagestyle{empty}\fi
% TODO: uncomment above line for final

\begin{abstract}
Object pose estimation is an essential computer vision problem in many robot systems. It is usually approached by estimating a single pose with an associated score, however, a score conveys only little information about uncertainty, making it difficult for downstream manipulation tasks to assess risk. In contrast to pose scores, pose distributions could be used in probabilistic frameworks, allowing downstream tasks to make more informed decisions and ultimately increase system reliability. Pose distributions can have arbitrary complexity which motivates unparameterized distributions, however, until now they have been limited to rotation estimation on \SOT{} due to the difficulty in training on and normalizing over \SET{}. We propose a novel method, SpyroPose, for pose distribution estimation using an \SET{} pyramid: A hierarchical grid with increasing resolution at deeper levels. The pyramid enables efficient training through importance sampling and real time inference by sparse evaluation. SpyroPose is state-of-the-art on \SOT{} distribution estimation, and to the best of our knowledge, we provide the first quantitative results on \SET{} distribution estimation. Pose distributions also open new opportunities for sensor-fusion, and we show a simple multi-view extension of SpyroPose. 
Project page at \href{http://spyropose.github.io}{spyropose.github.io}
\end{abstract}
\section{Introduction}
%Computer vision based pose estimation of rigid objects is an essential component of many robotics systems, where knowledge of an object's rotation and translation, referred to as the object's 6D pose, is required. Real world scenarios often involve relatively unconstrained environments with objects lying cluttered on a table or in a bin, making it a both challenging and crucial vision problem to solve. It is, therefore, not surprising that the topic is well studied in the literature. 
Many tasks in robotics involve manipulation of rigid objects and require that objects' rotations and translations, referred to as the objects' poses, are known. 
Vision systems are often relied upon to estimate object poses when the environment is relatively uncontrolled, such as when objects lie cluttered on a table or in a bin. 
%Since pose estimation is such a crucial but also challenging task, it is not surprising that the topic is well studied in the literature.

\begin{figure}
    \centering
    \includegraphics[width=0.9\linewidth]{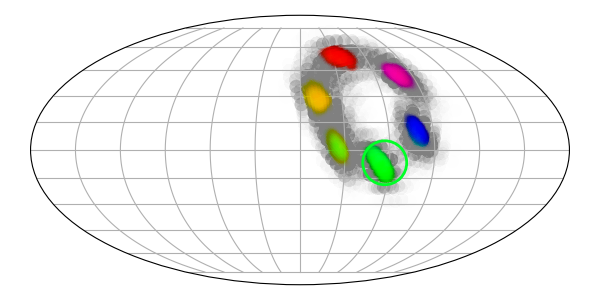}\\%
    \includegraphics[width=\linewidth]{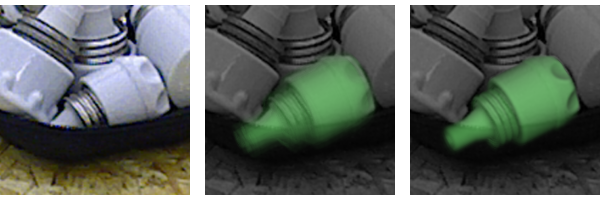}%
    \caption{%
        Visualization of \SET{} distributions at different levels of resolution in the pyramid.
        Bottom: Input image (left) and
        renders in green of poses weighted by their estimated probabilities for pyramid level three (middle) and five (right).
        Top: Marginalized \SOT{} distribution with two dimensions shown by a Mollweide projection and the last dimension by hue.
        To show both resolution levels in the same plot, level three is shown in grayscale.
        The true rotation is indicated by a circle. %The model correctly captures the visual ambiguity.
    }
    \label{fig:fig1}
    \figspacing{}
\end{figure}

Most of the pose estimation literature has been dedicated to algorithms which provide a single best guess of the pose.
Estimating a single pose can be adequate if the estimate is always good enough for the downstream task to succeed, or if the task is allowed to fail sometimes.
However, 
even with an ideal method,
there may be too much inherent visual ambiguity in an image to reliably perform the task, 
and for some tasks, failing is detrimental.

%there are also many real world tasks that involve visual ambiguity, e.g. due to occlusions or object symmetries. When pose estimation is confined to a point estimate, the vision system is by design limited in its capabilities to express uncertainty.

%Unlike point estimates, pose distribution estimation seeks to estimate an entire probability distribution over object poses. 
Representing uncertainties from vision can facilitate not only sensor-fusion, but also more informed, principled interactions between computer vision and robotics. 
This has e.g. been shown in \cite{hagelskjaer2019combined}, where the vision uncertainty, assumed to be normal, and success thresholds on a grasping task are combined to estimate the success of a grasp in a constrained environment. 
However, visual uncertainties are not always independent and normal, 
and making these assumptions inhibits probabilistic frameworks 
at the intersection of robot control and computer vision.
%research into more subtle interactions between robotic control and visual uncertainties 
%has been limited due to the lack of a reliable method for estimating complex pose distributions.% on \SET{}.

There are many ways to model probability distributions, including mixture models of parameterized functions, histograms, weighted ensembles of particles, and implicit functions. 
Implicit functions are especially interesting due to their ability to express arbitrary distributions
and they have been used successfully to estimate distributions on \SOT{} \cite{murphy2021implicit} 
using a contrastive loss with uniformly sampled negatives during training and a uniform grid during inference.
Due to the curse of dimensionality, there are two problems with extending this method to \SET{}.
Firstly, uniformly sampled negatives provide less information as the space grows from \SOT{} to \SET{}, 
and secondly, evaluation of a uniform grid on \SET{} becomes prohibitively expensive at any practical resolution.

We present SpyroPose, a novel method for pose distribution estimation,
using an \SET{} pyramid, a hierarchical grid, 
with higher resolution at deeper levels of the pyramid.
An example of an estimated pose distribution
under a six-fold symmetry ambiguity
is shown in Fig.~\ref{fig:fig1}.

Our method is based on three key ideas:
\begin{itemize}
    \item Importance sampling during training, enabled by the pyramid, providing harder negatives and lower variance estimates of the partition function.
    \item Sparse evaluation of the pyramid at inference, reducing the number of required evaluations by several orders of magnitude, enabling real time pose distribution estimation, even on a CPU. 
    \item Keypoint feature extraction, inducing a camera model bias into the model to enable translational equivariance and to avoid relying on a single latent embedding to represent complex, high-resolution distributions.
\end{itemize}
We present state-of-the-art rotation distribution estimation results on SYMSOL~\cite{murphy2021implicit} and TLESS~\cite{tless}, 
and to the best of our knowledge, we present the first quantitative results on \SET{} distribution estimation.
\section{Related Work}
Handling visual ambiguities is a challenging part of object pose estimation. 
Most work \cite{xiang2018posecnn, rad2017bb8, peng2019pvnet, labbe2020cosypose} on pose estimation uses manual annotations to explicitly handle one cause of visual ambiguity, symmetries, and infers a \enquote{best guess} pose estimate without expressing uncertainty.
The defacto pose estimation benchmark, BOP~\cite{hodavn2020bop}, is also centered around point estimates and known object symmetries.

Some methods \cite{hodan2020epos, surfemb} that provide point estimates do however handle visual ambiguities in a principled way.
\mbox{EPOS~\cite{hodan2020epos}} regresses dense, per pixel, histograms of object surface regions to obtain 2D-3D correspondences and use them in a PnP-RANSAC framework to provide pose estimates without knowing about symmetries a priori.
\mbox{SurfEmb~\cite{surfemb}} extends this idea and estimates dense full 2D-3D correspondence distributions.
However, the dense correspondence histograms and -distributions have not yet shown to be useful to model pose distributions.

There are also pose estimation methods which address pose uncertainties.
In \cite{shi2021fast}, model ensembles are used to estimate epistemic uncertainty which indicates a degree of generalization uncertainty caused by insufficient training data or a domain gap.
This however does not help represent the inherent aleatoric ambiguities in pose estimation.%, which could be used in down stream tasks or for sensor fusion.

Work on modeling the aleatoric uncertainty has traditionally been approached with parametric distributions.
The most common parametric model of uncertainty is the multivariate normal distribution.
Early work has focused on propagating correspondence uncertainties to pose space for classical pose estimation methods such as for ICP in \cite{censi2007accurate} and for PnP in \cite{franaszek2017propagation}. They assume that the correspondence ambiguities are independent and Gaussian, however, even simple symmetries and occlusions can break this assumption.

Rotation and position are defined on different manifolds, \SOT{} and $\R^3$, respectively, and it has been custom to assume them to be decoupled and treat the two parts separately, usually assuming the position to be normal.
For rotational uncertainty, Bingham is the most popular parametric distribution model.
Estimating the parameters of a Bingham mixture model has been done in various ways, 
    including deep learning based regression~\cite{okorn2020learning, deng2022deep, gilitschenski2019deep} 
    and fitting a Bingham mixture model to an ensemble of pose hypotheses~\cite{manhardt2019explaining}. 
Additionally, \cite{prokudin2018deep} shows that the parameters of a von Mises mixture model can be estimated.

Finally, there are works on unparametric distributions on \SOT{}.
\cite{okorn2020learning, kehl2017ssd6d} directly regress a rotation histogram which is able to represent arbitrary distributions at a coarse resolution, but the generalization across bins is limited, 
requiring all bins to be well represented in the training data, making it difficult to scale the histogram to higher resolution, not to mention \SET{}. 
\cite{sundermeyer2018implicit, deng2021poserbpf} trains a denoising autoencoder on image crops and uses cosine similarities on the learnt latent embeddings to estimate visual ambiguities. 
However, since the loss is a reconstruction loss rather than a probabilistic loss, and the distribution is designed rather than learnt, the resulting distributions are heuristic.

Recently, a number of methods has been proposed which model the unparametric distributions implicitly.
In \cite{kipode2022iversen}, the marginal distributions of keypoint projections are estimated across the image which can provide a conservative estimate of unnormalized pose likelihoods, 
however, the capacity of the model is limited because the joint distribution of keypoints is not modeled.
Their method could also be used for pose distribution estimation, however, they only present results on rotation distribution estimation due to the lack of a method to normalize over SE(3).
ImplicitPDF~\cite{murphy2021implicit}, which has inspired this work, trains a Multi Layer Perceptron (MLP) to map an image embedding and a rotation to an unnormalized likelihood.
They show a single qualitative result on \SET{} distribution estimation on simple synthetic data in their supplementary material, however, no quantitative results nor results on real data are shown.
HyperPosePDF~\cite{hofer2023hyperposepdf} is similar, but instead of estimating a latent embedding from an image and feeding it to an MLP, they learn a mapping from an image to the weights of an MLP which maps rotations to unnormalized likelihoods.
I2S~\cite{klee2023image} is also similar, but maps features from the image domain to \SOT{} followed by \SOT{} equivariant layers, before mapping to the unnormalized rotation likelihood.
They present good generalization capabilities but lower resolution than the related methods.

Like \cite{murphy2021implicit, hofer2023hyperposepdf, klee2023image}, we also use an implicit formulation to estimate unnormalized log likelihoods, however,
we use an \SET{} pyramid, a hierarchical grid, which enables importance sampling during training, allowing efficient learning of unparameterized \SET{} distributions.
The same pyramid is used at inference for sparse evaluation to enable real time pose distribution estimation.
We also make use of the spatial dimensions of the image in our latent pose distribution embedding, 
to relieve a single image embedding to represent complex, high-resolution distributions.
\section{Methods}
At its core, our method is based on learning pose distributions at different levels of resolution.
Given a pose hypothesis, we project object keypoints into the image, extract image features at the projected points and feed the sampled features to resolution-specific MLPs.
At inference, this allows a sparse top-down evaluation of a pyramid of poses, only expanding the most likely poses to the next, higher-resolution level.
During training, having models at different resolutions allows sampling negatives with known probabilities from a pose distribution which is closer to the model distribution than uniform, enabling importance sampling. %to provide better signal at higher-resolution layers.
%The following provides details of our method.

\subsection{Problem Definition}
Given an image crop, $I \in \R^{H \times W \times C}$, of an object of interest, we aim to estimate an unparameterized distribution, $p(x|I)$, of the object's six-dimensional pose, $x \in \SET{}$.

%The distribution should be unparameterized, i.e., not assuming position and orientation to be decoupled, and without assuming the distribution to be of a certain family or with a certain number of modes.
%The distributions should be normalized, and to have most practical value, the distributions should also both be of high resolution and be obtained in real time.

\subsection{\SET{} Pyramid Definition}
\label{sec:pyramid-def}
We use an equivolumetric hierarchical grid in \SET{}, which we'll refer to as an \SET{} pyramid.
Each layer of the pyramid is the cartesian product between a positional grid in $\R^3$ and rotational grid in \SOT{}.

For the rotational part of the pyramid, we use the HealPix grid \cite{gorski2005healpix} extended to \SOT{} by \cite{yershova2010generating}.
Like previous work \cite{murphy2021implicit, hofer2023hyperposepdf, kipode2022iversen}, we use the grid for its
equivolumetric property, but we also use its hierarchical structure.
Let $R^{(r)} \subset \SOT{}$ denote the grid of rotations at recursion $r$, and let $R^{(r)}_i \in R^{(r)}$ denote a cell in the grid, represented by its center.
The coarsest level, level 0, consists of 72 cells, $|R^{(0)}| = 72$, and
for each recursion, each cell is split into eight cells,
$|R^{(r)}| = 72 \cdot 8^r$.
The volume of the grid is $V(R^{(r)}) = \pi^2$ and because the grid is equivolumetric, a rotation cell has volume $V(R^{(r)}_i) = \pi^2 / (72 \cdot 8^r)$.

For the positional part, the bounds of the grid need to be defined, since $\R^3$ is unbounded. We define the bounds in two steps. 
In the first step, the positional error is modeled using a conservative estimate of visual ambiguity to ensure that the true pose is within the estimated bounds. In the second step, a hierarchical grid is defined such that it fully encompasses the conservatively estimated bounds.
Let $\hat t \in \R^{3 \times 1}$ be a coarse estimate of the object's position, $t$. 
Depending on the application, this estimate could come from a detector, be known a priori, or obtained otherwise. In this work, we assume $\hat t$ to come from a detector.
We then define a convservative bound on $t$ based on $\hat t$. Specifically, we presume that the maximum perceived positional ambiguity is equal to the object's radius.
Let $d$ be the diameter of the object of interest.
Then parallel to the image plane, we let the error be up to $d / 2$.
Along the view direction, we let the object's distance to the camera be down to half, which assuming a pinhole camera model would cause the appeared size to be approximately doubled.

Formally, we define a bound which meets the above criteria by introducing a truncated multivariate normal variable, $\tilde e = \mathcal{N}(0, \sigma I, 1/2)$, where "1/2" indicates, that its truncated at $||\tilde e||_2 = 1/2$, and then define our random error variable $e$ as
\begin{equation}
\label{eq:t-error-def}
e = A \tilde e, \quad
    A = \begin{bmatrix}
    d & 0 & \hat t_x \\
    0 & d & \hat t_y \\
    0 & 0 & \hat t_z \\
    \end{bmatrix}.
\end{equation}
The resulting bound of $\tilde t = \hat t + e$ is a sphere centered around $\hat t$ and elongated along the view direction, encompassing more depth- than in-plane ambiguity.
We represent a positional cell by its center,
and define the positional grid, $p^{(r)} \in \R^{3 \times N}$, with $N = 8^r$ at recursion $r$ as
\begin{equation}
\label{eq:pos-grid}
    p^{(r)} = \hat t + A g^{(r)}, 
\end{equation}
where $g^{(r)} \in \mathbb{R}^{3 \times N}$ consists of the centers of the cubes in the $2^r$ by $2^r$ by $2^r$ regular grid inside an origo-centered unit cube.
Note that $p^{(r)}$ encompasses $\tilde t$, since $g^{(r)}$ encompasses $\tilde e$. Also note, that $p^{(r)}$ is hieararchical and equivolumetric, since $g^{(r)}$ is hieararchical and equivolumetric and $g^{(r)} \mapsto \hat t + A g^{(r)}$ is an affine transformation.
The volume of the grid is $ V(p^{(r)}) = \det(A)$, since the volume of $g^{(r)}$ is 1. 
The volume of a cell in $p^{(r)}$ is thus $V(p^{(r)}_i) = \det(A)/8^r$.

The \SET{} pyramid is simply the cartesian product of the positional and rotational grid, however, it must be chosen at which recursion to align them.
At $R^{(0)}$, the angular distance to the nearest neighbour is approximately $\phi = 1\ \text{rad}$, causing a visual distance of up to approximately $d/2$.
Since $p^{(1)}$ has the same visual resolution, we define recursion 0 of the \SET{} pyramid, $x^{(0)}$, to be the cartesian product of $R^{(0)}$ and $p^{(1)}$: $x^{(0)} = R^{(0)} \times p^{(1)}$. 
Since $x^{(r)}$ is also an equivolumetric grid, it follows that
\begin{equation}
    V(x^{(r)}_i) = 
    \frac{V(x^{(r)})}{|x^{(r)}|} = 
    \frac{\det(A) \pi^2}{(72 + 8) 64^r}.
\end{equation}

To prevent our models from learning the structure of a fixed \SET{} grid, 
we randomly offset and rotate $p^{(1)}$ and rotate $R^{(0)}$ during training.
%For the rotational grid, we simply rotate $R^{(0)}$ by a random rotation during training, and for the positional grid, we let $p^{(r)} = \hat t + AR (g^{(r)} + s)$, where $s \in \R^{3\times1}$ and $R \in \SOT{}$, making sure to extend $g^{(r)}$ to make $p^{(r)}$ encompass $\tilde t$.

\subsection{Contrastive Loss}
The InfoNCE loss was presented in \cite{oord2018representation}, inspired by Noise Contrastive Estimation, 
\begin{equation}
\label{eq:infonce}
L_\text{InfoNCE} =  -\underset{x, I, X}{\E} \left[ \log \frac{
        f_\theta(x, I)
    }{
        f_\theta(x, I) + \sum_{x_j \in X} f_\theta(x_j, I)
    } \right],
\end{equation}
where $(x, I)$ is sampled from the data distribution $p(x, I)$,
$X$ is a set of $N$ samples from a noise distribution, $x_j \sim p_n(x)$,
and $\theta$ are the parameters of the model.
They show that for any $N$, the loss leads to approximating 
$f_\theta(x, I) \propto p(x | I) p_n(x)^{-1}$,
and it follows that letting the noise distribution be uniform, $f_\theta$ approximates an unnormalized distribution,
$f_\theta(x | I) = \tilde p(x | I; \theta) \propto p(x | I)$.

Note that the last term in the denominator of Eq.~\eqref{eq:infonce},
%\begin{equation}
$
    \hat Z = \sum_{x_j \in X} f_\theta(x_j, I),
$
%\end{equation}
is proportional to an unbiased estimate of the partition function,
$\E_X \; \hat Z \propto \int f_\theta(x, I) dx$.
An inherent problem with scaling the noise contrastive loss with uniform sampling of negatives to higher dimensions is that the variance of the partition function estimate becomes higher, due to the curse of dimensionality.

Importance sampling could be used to lower the variance of the estimate of the partition function, but it requires a heavy-tailed distribution close to $p(x | I; \theta)$ which can be sampled from with known sample likelihoods, and such a distribution is generally not available.
%However, in the next section we show how learning distributions at different resolutions can enable importance sampling.

\begin{table*}
    \small
    \centering
    \caption{%
        Rotation distribution estimation results on SYMSOL. 
        The table entries are estimated log likelihoods \textuparrow{} of the true rotation averaged over 5 k test images per object.
        Results below the gray line is on our implementation of SYMSOL I.
        For verification, we show \OursNoKP{} for both the original and our implementation of the dataset. 
        IS: Importance Sampling. KP: Keypoints.
    }
    \label{tab:symsol}
    \resizebox{\linewidth}{!}{%
    \begin{tabular}{lrrrrrrcrrrr}%
        \hspace{10em} &  
        \multicolumn{6}{c}{SYMSOL I}& 
        \hspace{1em} & 
        \multicolumn{4}{c}{SYMSOL II}\\
        \cmidrule{2-7}
        \cmidrule{9-12}
        Method \hspace{10em} & avg. & cone & cyl. & tet. & cube & ico. & 
               & avg. & sphX & cylO & tetX \\ 
        \toprule
        Prokudin et al. \cite{prokudin2018deep} (2018)
            &-1.87 &-3.34 &-1.28 &-1.86 &-0.50 & -2.39 & & 0.48 & -4.19 & 4.16 & 1.48\\
        Gilitschenski et al. \cite{gilitschenski2019deep} (2019) 
            & -0.43 & 3.84 & 0.88 & -2.29 & -2.29 & -2.29 & & 3.70 & 3.32 & 4.88 & 2.90\\
        Deng et al. \cite{deng2022deep} (2020) 
            & -1.48 & 0.16 & -0.95 & 0.27 & -4.44 & -2.45 & & 2.57 & 1.12 & 2.99 & 3.61\\
        ImplicitPDF \cite{murphy2021implicit} (2021) 
            & 4.10 & 4.45 & 4.26 & 5.70 & 4.81 & 1.28 & 
            & 7.57 & 7.30 & 6.91 & 8.49 \\
        I2S \cite{klee2023image} (2023) 
            & 3.41 & 3.75 & 3.10 & 4.78 & 3.27 & 2.15 &
            & 4.84 & 3.74 & 5.18 & 5.61 \\
        HyperPosePDF \cite{hofer2023hyperposepdf} (2023) 
            & 5.78 & 5.74 & 4.73 & 7.04 & 6.77 & 5.10 & & 7.72 & 7.73 & 7.12 & 8.53 \\
        \OursNoKPIS{} & 5.65 & 6.77 & 6.07 & 6.04 & 6.23 & 3.12 & 
            & 7.16 & 6.96 & 7.59 & 6.92 \\
        \OursNoKP{} & \textbf{7.33} & \textbf{7.62} & \textbf{6.46} & \textbf{8.69} & \textbf{8.63} & \textbf{5.23} & 
            & \textbf{9.27} & \textbf{9.07} & \textbf{9.32} & \textbf{9.41}\\
        \arrayrulecolor{gray}
        \midrule
        \arrayrulecolor{black}
        \OursNoKP{} & 7.12 & 7.37 & 6.54 & 8.39 & 8.62 & 4.70 \\
        \OursNoIS{} & 8.19 & 7.40 & 6.69 & 10.04 & 8.82 & 7.99 \\
        \Ours{} w/ cube KP & 8.86 & 7.55 & \textbf{7.06} & 9.60 & \textbf{10.58} & \textbf{9.50} \\
        \Ours{} & \textbf{8.97} & \textbf{7.67} & 6.96 & \textbf{11.04} & 10.10 & 9.06 \\
        \cmidrule[\heavyrulewidth]{1-7}
    \end{tabular}%
    }
    \figspacing{}
\end{table*}

\subsection{Pyramid Models \& Importance Sampling}
\label{sec:importance}
We use the loss in Eq.\eqref{eq:infonce} and let the positive sample be the cell in $x^{(r)}$ which encompasses the true pose, $x$.
Instead of learning $\tilde p(x|I; \theta)$ at one, high resolution, 
we learn the distribution at different resolutions,
one for each level in the pyramid, $\tilde \p(x^{(r)}_i |I; \theta^{(r)})$.
For a model at recursion $r$, the coarser models can then be used to provide an estimate of $\p(x^{(r)}_i |I; \theta^{(r)})$ which can be used for importance sampling.

Let $P(x^{(r)}_i) = x^{(r-1)}_{i \backslash 64}$ be the parent of $x^{(r)}_i$ in the previous recursion, where \textbackslash{} denotes integer division,
and let $C(x^{(r)}_i) = \left\{x^{(r+1)}_{64i+0}, \dots, x^{(r+1)}_{64i+63}\right\}$ be the
set of children of $x^{(r)}_i$ in the next recursion.
The siblings of $x^{(r)}_i$, including itself, is thus $S(x^{(r)}_i) = C(P(x^{(r)}_i))$, and let $S(x^{(0)}_i) = x^{(0)}$.

In the following notation, the parameters, $\theta^{(r)}$, and conditioning on $I$ is assumed and left out for clarity.
We denote the relative probabilities among siblings as
\newcommand{\q}{\text{q}}
\begin{equation}
    \q(x^{(r)}_i) = 
    \frac{
        \tilde \p(x^{(r)}_i)
    }{
        \sum_{x^{(r)}_j \in S(x^{(r)}_i)} \tilde \p(x^{(r)}_j)
    },
\end{equation}
and applying them recursively across resolutions results in a generative coarse-to-fine Markov chain model, similar to an auto-regressive model, but where the models are dedicated to resolutions rather than dimensions,
%A model at one recursion, $\p(x^{(r)})$, can be approximated by the coarser models, allowing efficient sampling with known probabilities by sampling a trajectory through the pyramid:
\begin{equation}
\label{eq:is-sampling}
    \p(x_i^{(r)}) \approx
    \bar \p(x^{(r)}_i) = 
        \q(x^{(0)}_{i \backslash 64^r})
        \q(x^{(1)}_{i \backslash 64^{r-1}})
        \cdots
        \q(x^{(r)}_{i}).
\end{equation}
%Note that $\sum_{x^{(r)}_i \in x^{(r)}} \bar \p(x^{(r)}_i) = 1$.
%
%Because we randomly offset the grids during training, there's an inherent ambiguity at the boundaries of the grid cells at each level,
%causing $\bar \p(x^{(r)}_i | I)$ to be a heavy-tailed approximation of $\p(x^{(r)}_i | I)$ at convergence.
%
In the InfoNCE loss in Eq.~\eqref{eq:infonce}, $\bar \p$ can thus be used as an importance sampling distribution,
\begin{equation}
    \hat Z_\text{IS} = 
        \sum_{x^{(r)}_i \in X} 
        \frac{
            \tilde \p(x^{(r)}_i)
        }{
            \bar \p(x^{(r)}_i)
        }, \quad
        x^{(r)}_i \sim \bar \p(x^{(r)}_i),
\end{equation}
which is proportional to an unbiased estimate of the partition function, like $\hat Z$, but with lower variance.%, assuming $\bar \p(x^{(r)}_i)$ is a better estimate of $\p(x^{(r)}_i)$ than uniform.
%
%We could sample from $\bar \p(x^{(r)}_i)$ independently at each recursion, but for efficiency, we use all the evaluated siblings at each recursion in $n$ trajectories through the whole pyramid,
%amounting to $|x^{(0)}|$ evaluations for the first layer and $64n$ evaluations for subsequent layers.

Note that we could maximize the log likelihood directly, normalized by the importance sampling partition function estimate, but initial experiments showed that the additional term in the denominator of Eq.~\eqref{eq:infonce} led to more stable training.
One intuitive reason is that if the negatives are easy, and $f_\theta(x, I)$ is the dominating term in the denominator of Eq.~\eqref{eq:infonce}, the gradient is close to zero, $\nabla_\theta L_\text{InfoNCE} \approx 0$. 
%so easy negatives cause a gradient close to zero. 
Without $f_\theta(x, I)$ in the denominator, this would not be the case.

\subsection{Inference with Pyramid Models}
At inference, only a sparse tree of the pyramid is evaluated, obtaining highest resolution where the probability is highest.
Initially, a distribution over all of the coarsest grid cells is obtained by the coarsest model, $\p(x^{(0)} | I; \theta^{(0)})$.
Let $x_k^{(0)}$ denote the top $k$ cells with respect to estimated probabilities, and let $p_k^{0}$ denote the sum of probabilities for $x_k^{(0)}$.
The top $k$ cells are then expanded to their children, $C(x_k^{(0)})$, which are evaluated by the next model to redistribute the probability $p_k^{0}$ with higher resolution:

\begin{equation}
    \hat\p(x_i^{(r)}) = 
        p_k^{(r-1)}
        \frac{
            \tilde\p(x_i^{(r)})
        }{
            \sum_{x_j^{(r)} \in C(x_k^{(r-1)})} \tilde\p(x_j^{(r)})
        }.
\end{equation}
This process is repeated until the last recursion, and the cells which have not been expanded, including all cells of the last recursion, are leaf nodes in the sparse tree.
The leaf node probabilities sum up to one and make up the estimated distribution.
The estimated likelihood of a continuous pose, $x$, is 
determined by the leaf node, $x^{(r)}_i$, encompassing $x$,
\begin{equation}
    \hat p(x) = \hat \p(x^{(r)}_i)V(x^{(r)}_i)^{-1}.
\end{equation}
%As long as a pose is within the estimated positional bounds, discussed in Section~\ref{sec:pyramid-def}, it will be represented by a leaf node.
%It is thus not detrimental if the true pose is not represented at the highest resolution, since it will be represented by a leaf node at a lower resolution.

%Note that when expanding poses at recursion $r$, the sum of the probabilities of the expanded poses in recursion $r+1$ must be equal to the sum of the parents' probabilities in recursion $r$, since the sum of all leaf nodes must sum to 1.

Note that in contrast to the importance sampling distribution from Eq.~\eqref{eq:is-sampling}, which is used during training, 
during inference,
relative probabilities among cells at a given recursion are entirely decided by the model at that recursion and are not affected by the probabilities at earlier recursions. 
This allows cell-border ambiguities at low resolution to be resolved at higher resolutions.

\begin{figure*}
    \centering%
    \begin{overpic}[width=\linewidth,unit=1mm,trim={0 24em 0 0},clip]{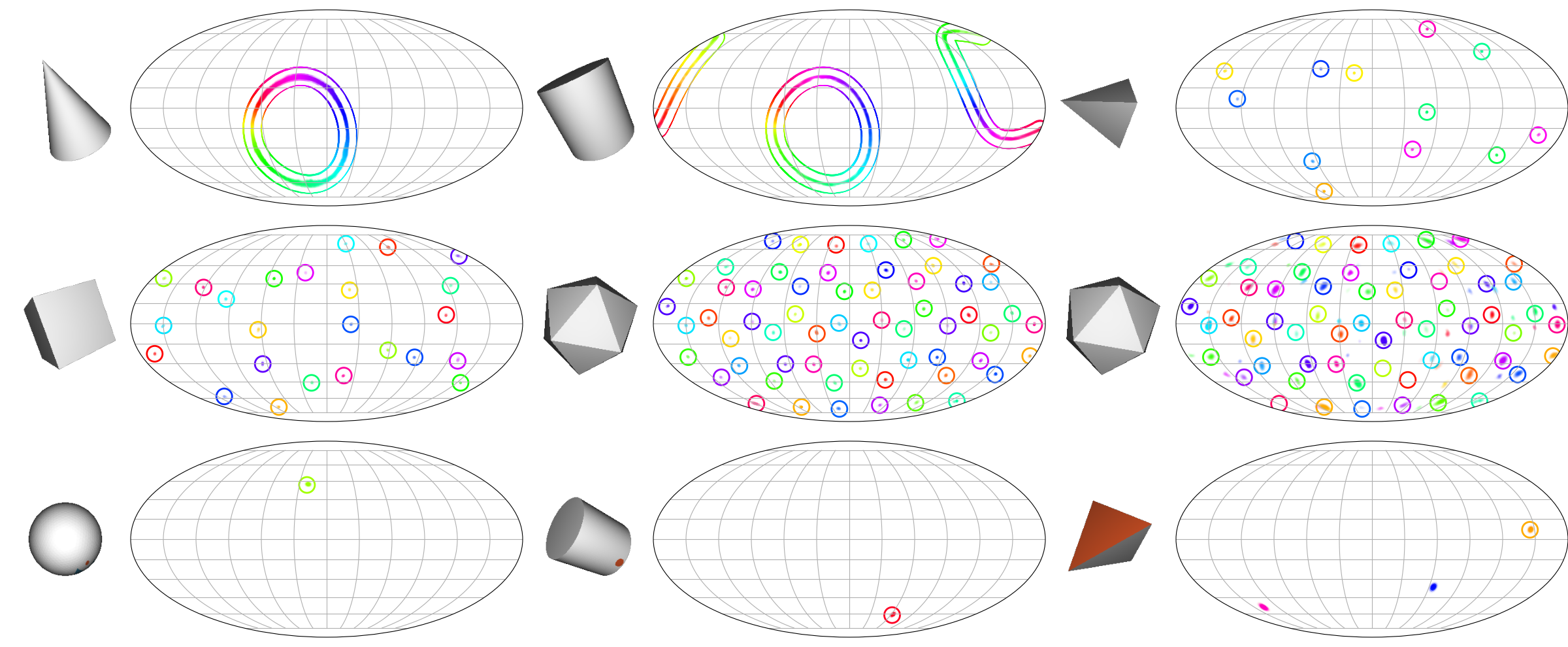}
    %    \put(70, 43){a)}
    %    \put(128, 43){b)}
        \put(70, 20){a)}
        \put(128, 20){b)}
    \end{overpic}
    \caption{%
        Qualitative SYMSOL I results. 
        We visualize the rotations at the last pyramid level (level 6) and their likelihoods as alpha, normalized for viewing.
        Circles, or for continuous symmetries donut-like shapes, indicate the correct rotation up to symmetry.
        a) and b) are from the same image, but b) shows our method w/o KP.
        Our method accurately captures all 60 modes of the icosahedron.
        %See supplementary for qualitative results on SYMSOL II.
    }
    \label{fig:symsol_dists}
    \figspacing{}
\end{figure*}

\subsection{Network architecture}
We use a UNet~\cite{ronneberger2015u} with a ResNet18~\cite{resnet} backbone to obtain a 64 dimensional feature map with the same spatial resolution as the image.
From the 3D mesh of the object, 16 approximately evenly spread keypoints are sampled with farthest point sampling.
Given a pose at the center of a pose cell, the keypoints are projected into the image and features are extracted from the feature map at the projected points with bilinear interpolation.
Keypoints that are projected outside the image receive a learnt out-of-image embedding.
The sampled keypoint features are concatenated and fed to a three-layer MLP with 256 hidden neurons. % and GELU activations.
The output of the network is a scalar, representing the estimated unnormalized log likelihood of the pose cell.
\section{Experiments}
We show results on \SOT{} distribution estimation, comparing with previous work, and then to the best of our knowledge, we show the first quantitative results on \SET{} distribution estimation.
We show \SOT{} results on SYMSOL I, SYMSOL II and TLESS;
and \SET{} results on TLESS and HB.
Lastly, we show a straight forward multi-view extension of SpyroPose which provides drastic improvements over single-view results, indicating the potential of sensor-fusion using unparameterized distributions.

\subsection{\SOT{} Results}
\paragraph{SYMSOL.}
ImplicitPDF~\cite{murphy2021implicit} introduced the synthetic dataset, Symmetric Solids (SYMSOL), for evaluation of distribution estimation on \SOT{}.
The dataset includes a variety of geometric primitives with different kinds of symmetries. 
The dataset has two parts:
SYMSOL I with texture-less objects, and SYMSOL II
%SYMSOL I consists of a cone, a cylinder, a tetrahedron, a cube and an icosahedron,
%and SYMSOL II consists of a sphere, a cylinder and a tetrahedron, 
with markers which are only visible in certain rotations, causing dynamic ambiguities.

The dataset does not include camera intrinsics or 3D models of the objects, and while the translation of the objects is fixed, the translation is unknown.
Since our method is based on projection of keypoints, we cannot apply it directly to the original SYMSOL dataset.
Instead, we apply a modified version of our method without keypoints on the original dataset, and implement SYMSOL I with known camera intrinsics, translation and 3D models, to evaluate our full method.

For our method without keypoints, we use a similar architecture as ImplicitPDF~\cite{murphy2021implicit}, with a ResNet50~\cite{resnet} to obtain a latent image embedding, a positional encoding of the rotation, and pass both embeddings to a small MLP. See \cite{murphy2021implicit} for details.
While the architecture of our method without keypoints is similar to ImplicitPDF,
our method still has an MLP for each level in the \SOT{} pyramid, and results are shown with and without importance sampling.
For our implementation of SYMSOL I, we approximately match perspective, scale and shader of the original dataset.
Since SYMSOL II has textures, it is not as easy to implement for a fair comparison. 

In our full method, we sample keypoints with farthest point sampling from the 3D model, however in some applications, 3D models may not be available, so we also show results where keypoints are chosen at the corners of two cubes with side lengths of 1 and 0.5 of the object diameter.

We train object-specific models 
with seven MLPs ranging from $R^{(0)}$ to $R^{(6)}$, and 1024 negatives per recursion per image.
For the models with importance sampling this is obtained with 128 sample trajectories, which with siblings at each recursion amounts to 1024 per MLP, because the branching factor is eight in the rotation pyramid. 
See Section~\ref{sec:pyramid-def} and \ref{sec:importance}.
%
% TODO: potentially remove below for space conciderations
Note that while the training set contains multiple rotation annotations due to symmetry, only the first of the provided annotations per image is used during training, assuming no knowledge about the symmetries.
For our full method, we use a batch size of 4.
Because the models without keypoints are computationally cheaper, we use a batch size of 16 for those to obtain similar training times.
%We found it beneficial to use batchnorm in the MLPs for our models without keypoints. %, which we found beneficial, especially for \textit{tet.} and \textit{ico.}.
We train all our \SOT{} models for 50~k iterations, approximately 2 hours per object on a single RTX 2080.

The results are provided in Table~\ref{tab:symsol},
and qualitative examples are shown in Fig.~\ref{fig:symsol_dists}.
We provide state-of-the-art results on SYMSOL I and SYMSOL II across all objects.
Our method predicts 24 and 130 times higher likelihoods on average for the true rotation than HyperPosePDF~\cite{hofer2023hyperposepdf} and ImplicitPDF~\cite{murphy2021implicit}, respectively.
Sampling keypoints from the surface of objects assumes that 3D models are available, but our results with cube keypoints perform almost as well, getting rid of this assumption.

Using keypoints provides a big improvement.
For our models without keypoints, with architectures similar to ImplicitPDF, a non-spatial latent embedding from the vision model has to express complex and high-resolution distributions.
Extracting keypoint features allows the model to use the image-space as an intermediate representation of the distribution and obtain translational equivariance.
%In addition to translational equivariance, there are also shorter gradient pathways to the early layers in the UNet encoder, which may improve training dynamics.

Because the evaluation of the \SOT{} pyramid is sparse,
a distribution down to recursion 6 with 18.8~M rotations only requires 21~k evaluations with $k=512$, 
almost three orders of magnitude fewer evaluations than evaluating the full grid. 
This allows our method to be run in real time, even on CPU, where it runs faster than ImplicitPDF on GPU.
See Table~\ref{tab:performance-so3}.

\begin{table}
    \small
    \centering
    \caption{%
    Inference time comparison for a single image.
    %We use $k = 512$, as for the other results.
    For our method we use an Intel i9-9820X CPU and an Nvidia RTX 2080 GPU.
    Batching improves fps further, obtaining 241 fps on \SOT{} for Ours w/o KP on GPU.
    %Grid Size: number of cells in the last recursion of the grid.
    \#eval: number of function evaluations.
    %Dev.: device.
    %fps: frames per second.
    }
    \label{tab:performance-so3}
    \vspace{.5em}
    \resizebox{\linewidth}{!}{%
        \begin{tabular}{clrrrr}%
            Space & Method & grid size & \#eval & dev. & fps \\
            \toprule
            \multirow{5.6}{*}{\SOT{}}
            & ImplicitPDF & 2.3 M & 2.3 M & gpu & 2.4 \\
            \arrayrulecolor{gray}\cmidrule{2-6}\arrayrulecolor{black}
            & \multirow{2}{*}{Ours w/o KP} & \multirow{4}{*}{18.9 M} & \multirow{4}{*}{21 k} 
                & cpu & 16.9 \\
            & & & & gpu & 53.5 \\
            & \multirow{2}{*}{Ours} & & 
                & cpu & 3.3 \\
            & & & & gpu & 48.3 \\
            \midrule
            \multirow{2}{*}{\SET{}} & \multirow{2}{*}{Ours}
            & \multirow{2}{*}{618 B} & \multirow{2}{*}{164 k} 
                & cpu & 0.5 \\
            &&& & gpu & 16.1 \\
            \bottomrule
        \end{tabular}%
    }
    \figspacing{}
\end{table}

While the pyramid allows us to efficiently evaluate beyond recursion 6, we are also SOTA if we evaluate at recursion 5 as ImplicitPDF, with avg. log likelihoods at 8.58, 8.42 and 8.80 for sphX, cylO and tetX, respectively.

\begin{figure}
    \centering
    \includegraphics[width=\linewidth, trim={0 0 -.5cm 0}, clip]{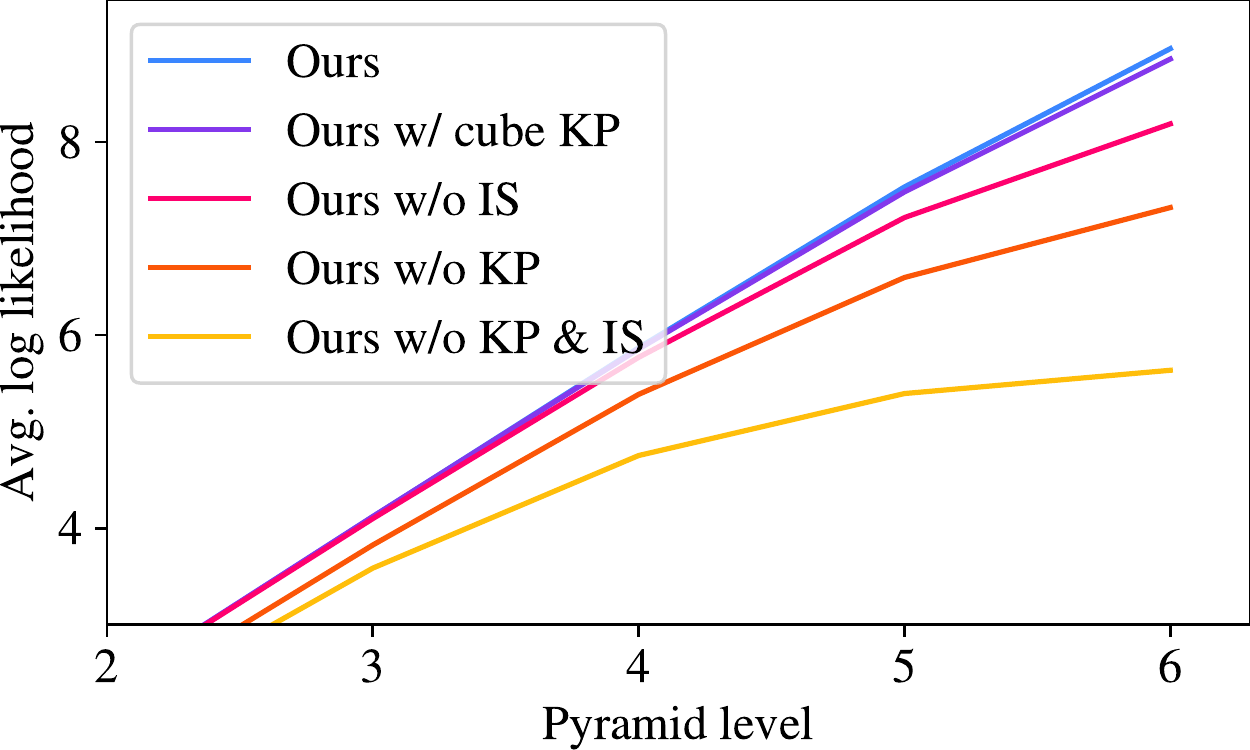}
    \caption{
        Log likelihoods on SYMSOL I, averaged over objects, at different recursion levels of the pyramid.
        Both keypoints and importance sampling improves learning at deeper levels.
    }
    \label{fig:symsol1-recursions}
    \figspacing{}
\end{figure}

Both keypoints and importance sampling improves learning at deeper recursions. See Fig.~\ref{fig:symsol1-recursions}.
The log likelihoods without keypoints and importance sampling flatten out around recursion 5, however, our full method could presumably benefit from even more recursions. %which is possible because of the sparse evaluation.

ImplicitPDF and HyperPosePDF train one model across all SYMSOL I objects, while we train a model per object, however, we use fewer function evaluations per object during training, and we provide similar improvements on SYMSOL II, where they also train a model per object.

\paragraph{Generalization.}
I2S~\cite{klee2023image} 
applies their method and ImplicitPDF's
method in a low-data regime,
training on SYMSOL with only 10k images instead of 45k.
I2S's ImplicitPDF models perform poorly and seem to have severely overfit,
so instead of re-reporting I2S's ImplicitPDF results, we note that Ours w/o KP \& IS is very similar to ImplicitPDF and performs similarly on the full dataset.
We train our models with and without importance sampling on 9.5k of the training images, using the remaining 500 for early stopping.
The results are shown in Table~\ref{tab:symsol-10k}. 
Our method outperforms I2S.
%The experiments indicate that importance sampling improves generalization quite significantly.
%It is unclear as to why. On the contrary, we would expect that higher variance in the gradient without importance sampling would have had a regularizing effect.

\begin{table}
    \normalsize
    \centering
    \caption{%
        Rotation distribution estimation on SYMSOL II 
        in a low-data-regime with 10k training images instead of 45k. LL \textuparrow{}.
    }
    \label{tab:symsol-10k}
    \vspace{.5em}
    \begin{tabular}{lrrrr}%
        Method & avg. & sphX & cylO & tetX \\
        \toprule
        %ImplicitPDF \cite{murphy2021implicit} & -0.73 & -2.51 & 2.02 & -1.70 \\
        I2S \cite{klee2023image} (2023) & 3.61 & 3.12 & 3.87 & 3.84 \\
        Ours w/o KP \& IS & 3.95 & 3.56 & 4.68 & 3.60  \\ 
        Ours w/o KP & \textbf{5.47} & \textbf{5.38} & \textbf{6.82} & \textbf{4.19} \\
        \bottomrule%
    \end{tabular}%
    \figspacing{}
\end{table}

\paragraph{TLESS.}
We follow ImplicitPDF and train a single model across all 30 objects.
Results are presented in Table \ref{tab:tless-so3}.
Our method provides almost an order of magnitude higher likelihoods than ImplicitPDF.
Note that recursion 5 of our method w/o KP \& IS provides an average log likelihood of 9.9, similar to ImplicitPDF.
This benchmark uses tight crops of the Kinect training images of singled-out objects,
uniform backgrounds and no occlusions,
for both training and testing using a random split.
The benchmark thus contains no domain gap, no need to generalize wrt. translation or handle occlusions or background clutter.
This is in contrast to the experiments in the following section.

\begin{table}
    \normalsize
    \centering
    \caption{
    Rotation distribution estimation on TLESS.
    %We present results for our method with and without importance sampling, for recursion level 5 and 6 (@R5, @R6).
    %\todo{psbly rm some lines}
    }
    \label{tab:tless-so3}
    \vspace{.5em}
    \begin{tabular}{lrr}
        Method & LL \textuparrow \\
        \toprule
        Prokudin et al. \cite{prokudin2018deep} (2018) & 8.8\\
        Gilitschenski et al. \cite{gilitschenski2019deep} (2019) & 6.9 \\
        %Deng et al. (2020) & 5.3 \\
        ImplicitPDF \cite{murphy2021implicit} (2021) & 9.8 \\
        %Ours w/o KP \& IS @R5 & 9.9 \\
        Ours w/o KP \& IS & 10.3 \\
        %Ours w/o KP @R5 & 10.9 \\
        Ours w/o KP & \textbf{11.9} \\
        \bottomrule
    \end{tabular}
\end{table}

\subsection{\SET{} results on TLESS and HB}
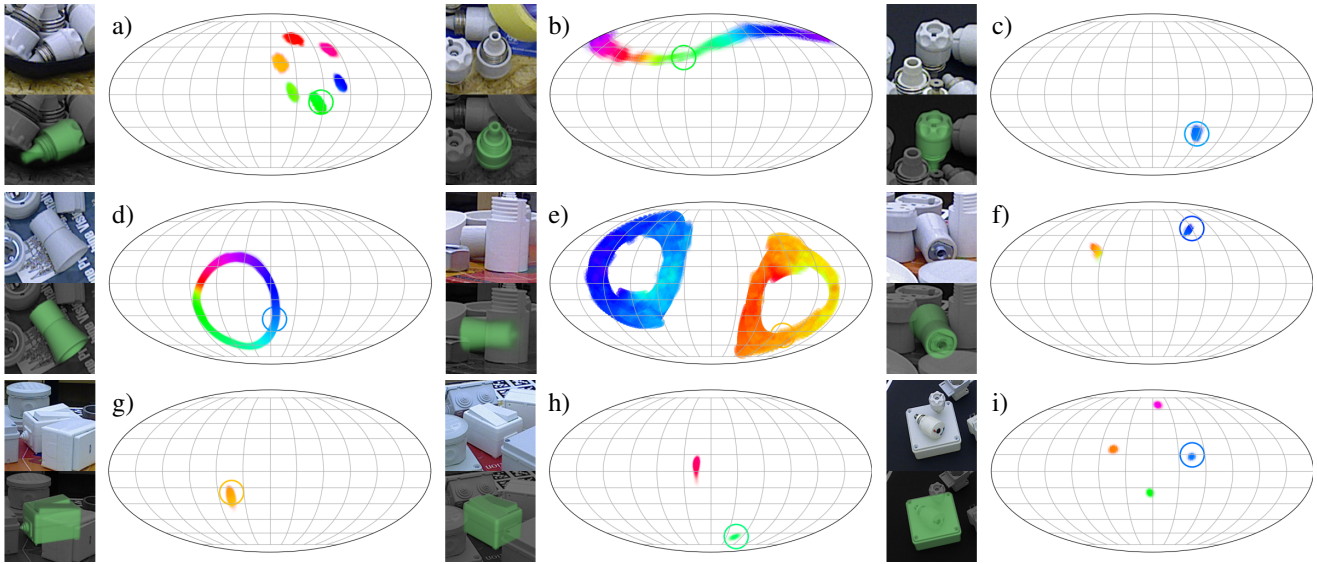
\begin{figure*}%
    \centering%
    \begin{overpic}[width=\linewidth,unit=1mm]{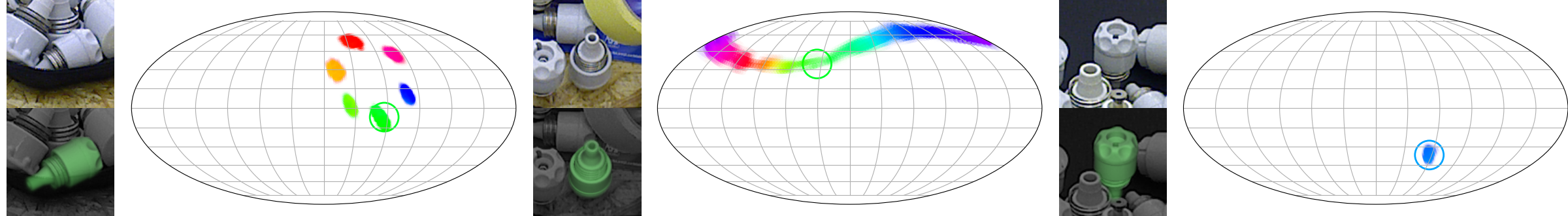}%
        \put(15, 20){a)}%
        \put(73, 20){b)}%
        \put(132, 20){c)}%
    \end{overpic}\\%
    \vspace{.2em}%
    \begin{overpic}[width=\linewidth,unit=1mm]{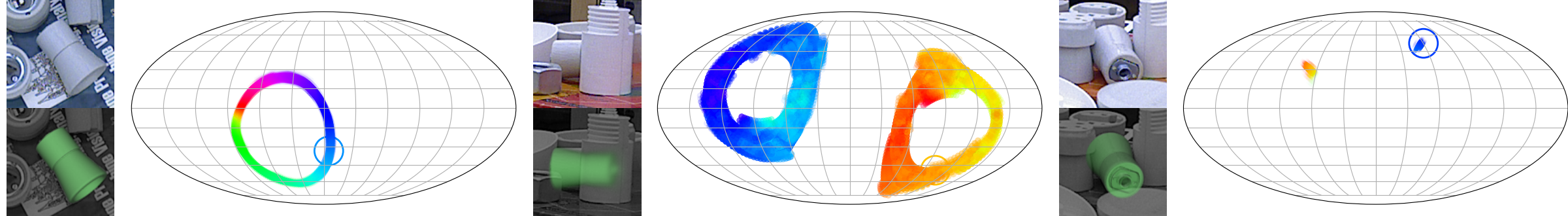}%
        \put(15, 20){d)}%
        \put(73, 20){e)}%
        \put(132, 20){f)}%
    \end{overpic}\\%
    \vspace{.2em}%
    \begin{overpic}[width=\linewidth,unit=1mm]{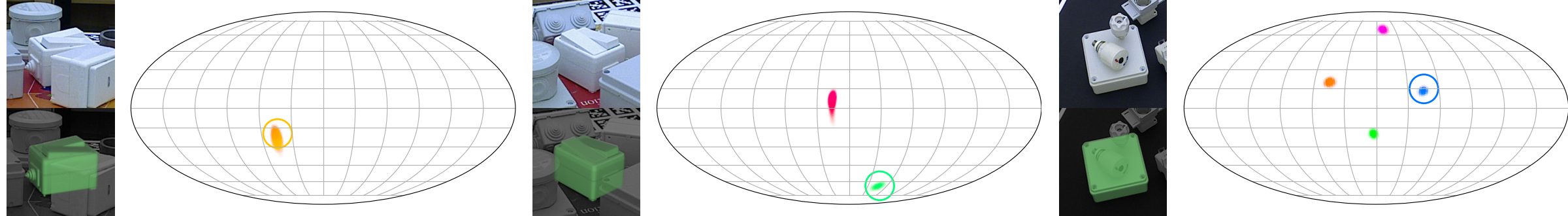}%
        \put(15, 20){g)}%
        \put(73, 20){h)}%
        \put(132, 20){i)}%
    \end{overpic}\\%
    %\begin{overpic}[width=\linewidth,unit=1mm]{media/tless_27_dist.png}%
    %\end{overpic}\\%
    \caption{%
        \SET{} distributions on TLESS.
        %The \SOT{} part of the distributions is visualized as previously.
        %To visualize the \SET{} distribution, we show distribution renders below the original images.
        First row shows distributions for object 1. 
        a) Six-fold rotational symmetry.
        b) Continuous rotational symmetry.
        c) No symmetry.
        Second row shows distributions for object 14. 
        d) Continuous rotational symmetry.
        e) The object of interest is behind the foreground object. Two-fold and continuous rotational symmetry. 
        Note that the two discrete modes have different depths, which can only be represented by a joint distribution.
        f) The continuous rotational symmetry is disambiguated by now visible features at the end of the object, and only a two-fold rotational symmetry along the same axis remains.
        g) and h) shows no symmetry and a two-fold rotational symmetry, respectively, for object 25.
        i) shows a four-fold rotational symmetry for object 27.
    }
    \label{fig:tless-dist}
    \figspacing{}
\end{figure*}

We train \SET{} pyramid models on TLESS~\cite{tless} and HB~\cite{kaskman2019homebreweddb} down to recursion level 5, 
selecting four objects from TLESS, representing different levels of symmetries,
and four objects from HB, which are common in the pose estimation community.
We train on the Physically Based Renders (PBR) from \cite{hodavn2020bop}
and show results on both held out PBR scenes as well as real images.
For TLESS, we show results on the real test images, and for HB, we show results on the real validation images, since test annotations are not publicly available. 
The results are presented in Table~\ref{tab:se3},
and qualitative examples on TLESS are shown in Fig.~\ref{fig:tless-dist}.

The chosen objects for TLESS have 16~k, 19~k, 23~k and 24~k image crops for training, 
and for HB: 25~k, 25~k, 24~k and 23~k.
Note that we have fewer images per object than in SYMSOL, 
there's a sim2real gap, 
and we're attempting to learn distributions on \SET{}.
For the above reasons, we regularize the models using dropout in the MLPs and heavy data augmentation during training. % including random rotation, color jitter, noise and blur.
%We also use dropout in the MLP as well as a keypoint dropout, which zeros an entire keypoint embedding across an image. Both dropouts have a 10\% chance.
We use 2048 negatives per recursion corresponding to 32 trajectories through the pyramid using Eq.~\eqref{eq:is-sampling}.

Since we are the first to present quantitative results on \SET{} distributions, 
there are no direct baselines to compare with, but we show results with and without importance sampling as well as results for a uniform distribution on the \SET{} grid. 
%Note that the latter changes slightly across objects, since the grid transformation $A$ depends on object size and depth (See Section~\ref{sec:pyramid-def}).
We considered comparing with point estimators, 
adding gaussians around point estimates,
but this could easily be worse than uniform,
since they would be punished severely for estimating a wrong mode.

We attempted to compare our joint distribution with a decoupled version,
but we were not able to successfully train a rotation model without keypoints, similar to ImplicitPDF, on these more challenging images. % (keypoint projection requires a joint distribution).
In fact, marginalizing SO(3) in our SE(3) distributions from our full method, we get 5.8 avg. log likelihood for the true rotation across the four T-LESS objects,
while w/o KP and w/o KP \& IS (similar to ImplicitPDF), we get 0.5 and 0.1, respectively. For comparison, -2.3 corresponds to a uniform distribution.
Learning joint distributions thus allows an architecture which significantly improves \textit{even} the marginal distributions on this more challenging benchmark.

On the held out PBR images, our model predicts 220 times higher likelihood of the true pose with importance sampling than without, compared to only 2 times higher on SYMSOL I.
This is consistent with Fig.~\ref{fig:symsol1-recursions},
which indicates that importance sampling becomes more beneficial when the probability is more concentrated, since evaluation of uniform samples then provides less information.
%
%On the real images, importance sampling is still beneficial, albeit not to the same degree.
%Some of this discrepancy is likely due to various domain gap effects such as inaccurate 3D models and inaccurate annotations.
%Despite the domain gap, our model predicts 25~M times higher likelihood for the annotated ground truth pose than a uniform distribution in the grid.

%We visualize the rotational part of the \SET{} as previously.
%To visualize the joint distribution over position and rotation, 
%we render leaf poses in the pyramid and weigh the renders with respect to their poses' estimated probabilities.

\begin{table*}
    \normalsize
    \centering
    \caption{%
        \SET{} distribution estimation on four representative objects from each of TLESS and HB. %symmetries as well as four HB objects.
        %The four HB objects are Wrench, Driller, Holepuncher and Phone, respectively.
        Models are trained on synthetic data (PBR), and we present results on both held out PBR images and real images.
        Entries are avg. log likelihoods of the ground truth poses.
        %We note that we have only trained models on these eight objects, so the results should be unbiased.
    }
    \label{tab:se3}
    \begin{tabular}{clrrrrrcrrrrrcrrrrr}
        & & \multicolumn{5}{c}{TLESS}
 %       & & \multicolumn{5}{c}{ITODD}
        & & \multicolumn{5}{c}{HB} \\
        \cmidrule{3-7}\cmidrule{9-13}\cmidrule{14-19}
        Data & Method 
            & avg. & 1 & 14 & 25 & 27 
            && avg. & 2 & 7 & 9 & 21 
            \\
        \toprule
        \multirow{3}{*}{PBR}
        & Uniform 
            & 2.7 & 3.5 & 3.2 & 2.4 & 1.7 &
            & 0.8 & 0.4 & 0.5 & 1.4 & 0.9 \\
        & Ours w/o IS 
            & 16.7 & 16.5 & 16.6 & 17.8 & 16.0 &
            & 15.9 & 16.3 & 15.1 & 16.4 & 15.8\\
        & Ours 
            & 21.6 & 22.4 & 20.5 & 22.6 & 20.9 &
            & 21.7 & 21.7 & 21.2 & 21.8 & 21.9\\
        \arrayrulecolor{gray}\midrule \arrayrulecolor{black}
        \multirow{3}{*}{Real}
        & Uniform 
            & 2.7 & 3.5 & 3.3 & 2.4 & 1.7 &
            & 0.9 & 0.6 & 0.6 & 1.5 & 1.0\\
        & Ours w/o IS
            & 16.6 & 14.1 & 15.9 & 19.2 & 17.2 &
            & 16.0 & 18.6 & 16.5 & 13.2 & 15.8\\
        & Ours 
            & 18.8 & 16.9 & 17.5 & 20.8 & 20.0 &
            & 18.9 & 21.3 & 20.6 & 16.4 & 17.3\\
        \bottomrule
    \end{tabular}
    \figspacing{}
\end{table*}

\begin{table}
    \normalsize
    \centering
    \caption{\SET{} distribution estimation results on TLESS. LL\textuparrow{}.}
    \label{tab:multiview}
    \vspace{0.2em}
    \resizebox{\linewidth}{!}{%
        \begin{tabular}{lrrrrr}
            Method & avg. & 1 & 14 & 25 & 27\\
            \toprule
            Ours & 18.8 & 16.9 & 17.5 & 20.8 & 20.0\\
            Ours w/ $A=dI$ & 18.4 & 16.4 & 16.6 & 20.9 & 19.9\\
            Ours w/ Multi-view & 25.2 & 23.7 & 23.6 & 28.2 & 25.2\\
            \bottomrule
        \end{tabular}
    }
    \figspacing{}%
    \vspace{-1em}
\end{table}

\subsection{Multi-view}
Lastly, we show a straight forward application: multi-view pose estimation. %, since our distributions lend themselves very naturally to such an extension.
Since uncertainty is represented in pose space rather than image space,
it is well-suited for principled sensor-fusion,
combining information from multiple sources. % in a principled way.
%\SET{} distributions have many potential applications.
%For example, we note that our method could also be used to learn action distributions in \SET{} for robotics.
With multi-view crops and known extrinsics, we let $A = d I$, such that $p^{(r)}$ is a cubic grid, and use the same grid in a common frame across views.
%Since our models have been trained with randomly offset grids, we can use the same single-view models.
For each recursion in the pyramid at inference, the unnormalized log likelihoods are simply averaged across views.
We show multi-view pose distribution estimation results on TLESS in Table~\ref{tab:multiview} with the same sets of up to four views as in \cite{labbe2020cosypose, haugaard2022multi}.
We note that $A = dI$ alone do not improve performance. % and provide the results for verification.
In fact, it can be harmful to have too much resolution along depth at inference, as most nodes are then spent on representing depth ambiguity.
This simple extension increases the likelihood of the true pose by almost three orders of magnitude.
\section{Conclusion}
This work has proposed SpyroPose, a novel method for pose distribution estimation on \SET{}. 
Our method is based on learning pose distributions at different levels of resolution using a hierarchical \SET{} grid, a pyramid,
which enables importance sampling for efficient learning at training time 
and sparse evaluation at inference, allowing real time pose distribution estimation. %, even on CPU.
Our method outperforms state-of-the-art methods for rotation distribution estimation on \SOT{} on the SYMSOL and TLESS datasets, and to the best of our knowledge, we provide the first quantitative results on pose distribution estimation on \SET{}.
Lastly, 
with a straight forward multi-view extension of SpyroPose,
we have shown how easily pose distributions allow information to be fused from multiple sources,
%provided an example application of our distributions, using them for multi-view pose estimation, 
showing great potential for our method as a core component of future work.
%Code will be made publicly available.

{\small
\bibliographystyle{ieee_fullname}
\bibliography{egbib}
}

\end{document}